# Optimal Trajectory Planning for Flexible Robots with Large Deformation

Sajjad Edalatzadeh[1]

## Abstract

Robot arms with lighter weight can reduce unnecessary energy consumption which is desirable in robotic industry. However, lightweight arms undergo undesirable elastic deformation. In this paper, the planar motion of a lightweight flexible arm is investigated. In order to obtain a precise mathematical model, the axial displacement and nonlinear curvature of flexible arm arising from large bending deformation is taken into consideration. An in-extensional condition, the axial displacement is related to transverse displacement of the flexible beam, is applied. This leads to a robotic model with three rigid modes and one elastic mode. The elastic mode depends on time and position. An Assume Mode Method is used to remove the spatial dependence. The governing equations is derived using Lagrange's method. The effects of nonlinear terms due to the large deformation, gravity, and tip-mass are considered. Control inputs include forces and moment exerted at the joint between slider and arm (see Fig. 1). The conventional computed torque control laws cannot stabilize the system, since there are not as many control inputs as states of the system. A Particle Swarm Optimization (PSO) technique is then used to obtain a suitable trajectory with the aim of minimizing excitations of the elastic mode. Two methods are considered for generating a trajectory function, either to use a three-layer Artificial Neural Network (ANN) or to use spline interpolation. A sliding mode control strategy is proposed in which the sliding surfaces include elastic mode in order to guarantee robustness. The simulations show that the three-layer ANN technique provides arbitrary small settling time, and also the optimization algorithm converges faster and generates smooth trajectories unlike spline function technique.

[1] msedalatzadeh@gmail.com

# 1. Introduction

Manipulators with flexible arms have extensive application in robotic industry due to higher length, greater payload-to-manipulator-weight, lower cost, fast operation, lower energy consumption, and better transportability. These are specifically important in space robotics applications [1]. However, the beam flexibility leads to undesirable arm vibrations. The vibrations are problematic in two ways: first, the modeling of the flexible arms are more challenging than that of rigid arms [2]; second, the control synthesis is challenging [3].

If the mathematical model of a flexible arm is inaccurate, the positioning of the end effector becomes inaccurate too [2]. several mathematical models have been suggested for flexible arms [4]. The simplest model is discrete spring-mass model (e.g., see [5]), which has a number of limitations. Not only does it neglect continuity of flexible arms, which characterized by infinite number of degrees of freedom, but also it is not clear how to determine a suitable spring constant. In continuous models of flexible arms, linear Euler-Bernoulli beam model has widely been used in the literature [6-9]. In this model, it is assumed that the shear deformation and the rotary inertia of cross-sectional element around lateral axis are negligible. This means normal to the natural surface remains normal throughout a deformation [10]. This is a relatively accurate model of the arm if the arm is short relative to its diameter; although such arms can also be considered as rigid links [6]. The response amplitude of slender beams can become large relative to its length. This is why non-linear terms arising from curvature and axial displacement begin to influence the dynamics [11]. Sathyamoorthy [12] surveyed studies on the effect of geometric nonlinearities arising from axial stretching and curvature in beams. In most studies on nonlinear beams, only nonlinear terms due to the axial shortening (also known as nonlinear inertia) is considered. However, Crespo de Saliva and Glynn [13, 14] showed that the nonlinear terms due to the curvature influence the flexural response of the beam the same order of magnitude as nonlinear inertia. Nayfeh and Pai [15] showed that the often-neglected geometric nonlinearities creates a hardening effect, but the nonlinear inertia creates a softening effect. They also showed that the dominant nonlinearity for the first mode is of the hardening type, whereas in the second and higher modes, the dominant nonlinearity is of softening type [16]. The authors also conducted a survey including a detailed derivation of the governing equations of different nonlinear beams (see ref. [16], chapter 4).

Over the years, researchers attempted various techniques to control manipulators with flexible beams [3, 17-19]. In particular, Spong et al. [20], Siciliano and Book [21], and Lewis and Vandegrift [22] applied singular perturbation method to control flexible link arms. In this method, the system dynamics are decomposed into a fast subsystem representing the elastic links motion and a slow subsystem representing the rigid body motion. The control law includes two terms with different time scales. The first term is based on computed torque method and is applied to slow dynamics in order to achieve joint trajectory tracking. The second term is included to influence the fast subsystem with the aim of suppressing residual vibrations of flexible links. The singular perturbation control technique is less effective if the fast and slow time scale are relatively close [23]. This is the case for large deformable flexible manipulators. In addition, for nonlinear under-actuated systems, finding a state feedback control law that guarantees global stability is challenging. Instead, a feed-forward control law could be computed to suppress the residual vibration of elastic modes (see ref. [3], conclusion section). Input shaping and trajectory planning are two well-known feed-forward techniques; Input shaping is widely used, and trajectory planning yields better result for highly nonlinear systems [24]. K. Park and Y. Park [25] designed an optimal trajectory for a two-link flexible manipulator combining Fourier series and polynomial functions. They showed that the optimal trajectory creates the least vibrations in the links. K. Park then used this method in a trajectory planning task in presence of obstacles [26] and under torque constraints [27]. Benosman [28] used the concept of trajectory planning to investigate the rest-to-rest motion of a multi-link planar flexible manipulator in a given fixed time. The method is based on backward integration of elastic dynamics. Abe proposed a novel trajectory planning technique to control the residual vibration of a large deformable flexible manipulator [29] and a flexible Cartesian manipulator [30] in a point-to-point movement. In order to generate suitable and smooth trajectories, Abe used a spline function in [29] and an artificial neural network in [30]. A particle swarm optimization (PSO) algorithm was later employed to find an optimal trajectory. It should be noted that in the presence of disturbance and model uncertainty, the feed-forward controllers might fail to completely suppress the residual vibration of flexible manipulators with small structural damping. A robust control law such as slide mode control law can resolve this issue. For instance, Yin et al. adopted an augmented sliding mode control law in [24]. The author also conducted an experiment to show that the controller can completely suppress the residual vibration, which would be inevitable without a robust controller.

The study of nonlinear beams is, for the most part, limited to static or dynamic response of either externally excited beams or parametric excited beams [16] in which the rigid body motion of the beam has been neglected. In addition, in most studies, linearized models of the link flexibility are utilized; this puts a limit on the size of link deformations. Dwivedy and Eberhard [2] conducted a comprehensive review of the literature on the dynamic analysis and control of flexible manipulators. The survey looks at 433 papers published between 1974-2005. In these papers linear beam model have been considered where large deformations are not permitted. Flexible manipulators with nonlinear model are rarely studied. Damaren and Sharf [31] studied the dynamics of two-link Space Shuttle Remote Manipulator System. They studied the effect of various types of nonlinearity including inertial and geometrical nonlinearities. The resulta show the inertial nonlinearity is necessary for accurate and fast maneuvers. Boyer et al. [32-34] generalized the Newton-Euler model for flexible manipulator by considering the nonlinear kinematic of Euler-Bernoulli beam. The author derived the governing equations of flexible robots that has several flexible links. Martins et al. [35] derived linear and nonlinear models for a flexible rotating manipulator by considering linear and quadratic approximation of the rotation matrix. They explored the effect of nonlinear terms due to axial shortening, shear deformation, and centrifugal force. They ran numerical simulation to show that without accounting axial shortening, the centrifugal softening term leads to instability, and by adding the axial shortening term the model will be stable. Yang et al. [36] studied the same manipulator without assuming that the beam neutral axis is inextensible; thus, the axial displacement is considered as an independent variable. The author derived a discretized model where the axial displacement is set to zero in order to investigate the vibration control of small transverse displacements. Piezoelectric sensors and actuators are bonded at the roof of the beam; and a combination of positive position and momentum exchange feedback control law utilized to suppress the transverse vibration of the rotating beam. Xue and Tang [37] studied the same rotating beam model, but assumed that the hub torque is an external uncontrolled input. An integral sliding mode control was proposed to suppress the beam vibrations due to hub rotation and external disturbances. The system without piezoelectric actuator is underactuated, but controllable. The control inputs on rigid modes could be designed such tracking and vibration suppression can be achieved concurrently. Al-Bedoor and Hadman [38] considered the nonlinearity that arises from higher order terms in the series expansion of curvature. Centrifugal force and axial displacement are not included in the potential energy of their beam model. Abe [29] studied a two-link rigid-flexible manipulator with tip mass, and expanded the nonlinear curvature to second order terms. The author proposed a novel trajectory tracking to suppress residual vibrations of elastic modes in a point-to-point motion. Fazel et al. [39] investigated the dynamic behavior of a rotating flexible beam, in which the nonlinear curvature and inextensional condition are taken into account. They employed two different numerical methods to solve the nonlinear differential equations.

It is evident from the existing literature that all types of nonlinearity, namely centrifugal force, axial shortening, and nonlinear curvature are not considered concurrently in the modeling. The gravitational force is often neglected; and the revolute joint in rotating flexible arms is often stationary. The primary goal of present work is to fill these gaps by including all the nonlinear terms into the model of rotating flexible arm. The effect of gravity and tip mass are also taken into consideration and the revolute joint is allowed to move in a plane. In addition, point-to-point trajectory planning of the manipulator is investigated by using a combination of trajectory planning and an augmented sliding mode control technique.

This paper is consequently divided into 8 sections. Section 2 presents a detailed derivation of the model. Section 3 introduces the concept of trajectory planning control scheme. In this section, two different methods of generating a trajectory was proposed. The PSO algorithm is employed to find an optimal trajectory. Section 4 presents the sliding mode control technique for trajectory tracking. Section 5 presents simulation results and provides a comparison of proposed control schemes and the effect of large deformation, gravity, and tip mass on the controllers.

## 2. Governing equation of flexible manipulator

Figure 1 shows a schematic of a planar single-link large deformable manipulator with a revolute and two prismatic joints. The frame $OXY$ denotes the inertial fixed frame with origin $O$ initially located at the center of a rigid hub. A reference moving frame $O'X'Y'$ is attached to the hub with the same center as of hub. Axis $O'X'$ is oriented along the neutral axis of the undeflected beam and rotates relative to inertia frame $OXY$ by angle $\theta(t)$. The beam has initial length $L$, flexural rigidity $EI$, and mass per unit length $\rho$; a mass $m$ attached at the tip of beam. The hub has radius $R_H$ and polar moment of inertia $J$; the mass of movable slider is denoted by $M$.

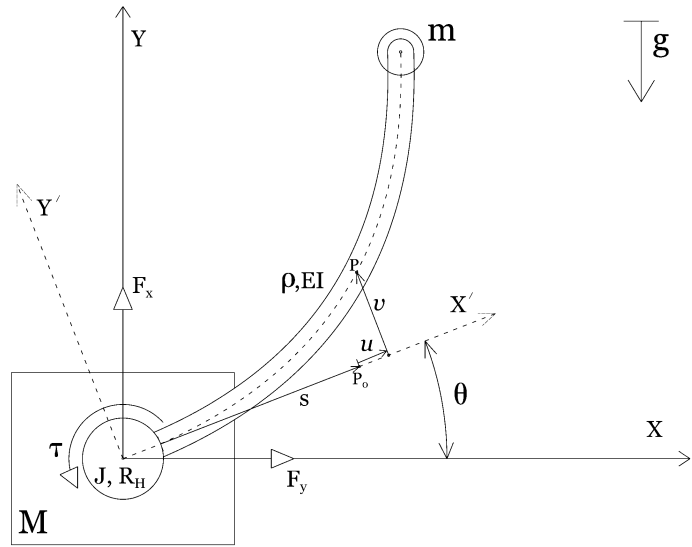

**Fig 1.** Schematic of rotating-sliding flexible arm that carry tip mass in a vertical plane.

A material point $P_o$ on the neutral axis has an initial distance $s$ form the root of the beam; under deformation, it moves along the $O'X'$ and $O'Y'$ axes by an amount of $u$ and $v$, respectively, to reach the point $P$. The position vector of point $P$ can be expressed as

$$\vec{r}(s,t) = (R_H + s + u)\hat{i} + v\hat{j}, \tag{1}$$

where $\hat{i}$, $\hat{j}$ denote unit vectors along $O'X'$ and $O'Y'$, respectively. The velocity of point $P$ with respect to the inertia frame can be derived by taking the time derivative of Eq. (1)

$$\dot{\vec{r}}(s,t) = (-\alpha\sin(\theta) + \beta\cos(\theta) + \dot{X})\hat{I} + (\beta\sin(\theta) + \alpha\cos(\theta) + \dot{Y})\hat{J}.$$
$$\begin{cases} \alpha = \dot{\theta}(R_H + s + u) + \dot{v}, \\ \beta = \dot{u} - \dot{\theta}v, \end{cases} \tag{2}$$

where dot denotes the derivative with respect to time; $\hat{I}$ and $\hat{J}$ are unit vectors of inertia fixed frame; $X$ and $Y$ denote the horizontal and vertical displacement of the slider; $u$ and $v$ represent the axial and transverse deflection of the beam, respectively. Subsequently, the kinetic energy of the manipulator can be written as

$$T = \frac{1}{2}\int_0^l \rho \dot{\vec{r}} \cdot \dot{\vec{r}}\, ds + \frac{1}{2}J\dot{\theta}^2 + \frac{1}{2}m\dot{\vec{r}}_m \cdot \dot{\vec{r}}_m + \frac{1}{2}M(\dot{X}^2 + \dot{Y}^2), \tag{3}$$

where $\dot{\vec{r}}_m = \dot{\vec{r}}(l,t)$ is the velocity of tip mass. In order to obtain a precise mathematical model, all terms that contribute to potential energy of the manipulator must be considered. Accordingly, the elastic potential energy arising from axial and transverse deflection of beam, the rotational potential energy induced by the coupling of axial centrifugal forces and bending deformations, and the gravitational potential energy of all masses are included in the model. That is,

$$U = \frac{1}{2}\int_0^l EI\kappa^2 ds + \frac{1}{2}\int_0^l EAu'^2 ds + \frac{1}{2}\int_0^l Fv' ds$$
$$+ \int_0^l \rho g\left[(s+u)\sin(\theta) + v\cos(\theta) + Y\right]ds + mg\left[(l + u(l,t))\sin(\theta) + v(l,t)\cos(\theta) + Y\right] + (M + m_H)gY, \tag{4}$$

where the first two terms are due to the elastic potential energy. Under the assumption of Euler-Bernoulli beam, the elastic potential energy of beam arising from the transverse deflection is correlated to the curvature of the beam [16]. The third term is due to the rotational potential energy, where $F$ denotes the centrifugal force and is calculated as

$$F(s) = \int_s^l \rho\dot{\theta}^2 (R_H + \zeta) \, d\zeta = \rho\dot{\theta}^2 [R_H(l-s) + \frac{1}{2}(l^2 - s^2)]. \quad (5)$$

The three last terms are gravitational potential energy of the beam, tip mass, and the collection of cart and hub, respectively. In addition, the virtual work $\delta W$, which is caused by the external non-conservative horizontal force $F_x$ and vertical force $F_y$ and torques $\tau$ exerted at hub, is

$$\delta W = F_x \delta X + F_y \delta Y + \tau \delta \theta. \quad (6)$$

Let $\phi$ be the angle of tangent lines to beam with $O'X'$ axis (i.e. $v' = \partial v/\partial s = \sin(\phi)$). Then, the curvature of beam is $\kappa = \partial \phi/\partial s = \phi'$; it can be expressed in terms of transvers deflections as [16]

$$\kappa^2 = \frac{v''^2}{1 - v'^2}. \quad (7)$$

Moreover, by assuming the beam to be inextensional, the axial deflection of beam can be correlated to transverse deflection of beam as well [16], That is

$$e = \frac{ds - ds^\circ}{ds^\circ} = \sqrt{(1+u')^2 + v'^2} - 1 = 0 \rightarrow \begin{cases} u(s,t) = -s + \int_0^s \sqrt{1-v'^2} \, ds, \\ \dot{u}(s,t) = \int_0^s v'\dot{v}/\sqrt{1-v'^2} \, ds, \end{cases} \quad (8)$$

where $e$ is axial strain, defined as ratio of the change in length to the original length. A clamped boundary condition at the root of the beam was considered in order to integrate $u'$ (i.e. $u$, $v$ and $v'$ are set to zero at $s = 0$). Inextensional assumption reduces the spatially dependent variables from two to one, and hence the position of flexible manipulator can be described by four independent variable namely, $v(s,t)$, $\theta(t)$, $X(t)$, and $Y(t)$. Since the model is intended to allow large flexural vibration, the nonlinear terms $1/(1-v'^2)$, $1/\sqrt{1-v'^2}$, and $\sqrt{1-v'^2}$ in Eqs. (7) and (8) should be approximated up second order terms. The nonlinear curvature and axial deflection consequently approximated as follows:

$$\begin{cases} u(s,t) \simeq -\dfrac{1}{2}\int_0^s v'^2\, ds, \\ \kappa^2 \simeq v''^2 + v''^2 v'^2. \end{cases} \qquad (9)$$

Finally, the original dynamic of flexible beams would be described by a Partial Differential Equation (PDE) in which states evolve in an infinite-dimensional space. For such mathematical models, control synthesis design methods are not readily available. Researchers often truncate the PDE models or use of the boundary control theory of PDEs. For example, Lagnese and Leugering [40] proposed a boundary control feedback law for a PDE governing a nonlinear cantilever beam. They proved well-posedness and stability of closed-loop system. Most commonly, the governing equations are truncated into a finite set of Ordinary Differential Equations (ODEs). This is often done using Finite Element Method (FEM) or Assume Mode Method (AMM). For instance, Bayo [41, 42] used FEM to discretize the governing equation of rotating flexible arm for a tip positioning task. Despite some advantages of FEM, the simulation time for large set of ODEs will be high and the control laws are often complicated. Also, such control laws sometimes need to feedback from internal nodes, i.e. inside the flexible beam region. This is not applicable in most industrial robots [43]. Moreover, over-estimated stiffness matrix calculated from FEM can possibly destabilize the original system under a model-based control algorithm [44]. On the other hand, Assumed Mode Method (AMM) requires less simulation time and yields better results for a single-link flexible manipulator with uniform cross-section [44]. For example, Al-Bedoor et al. [38, 45] studied the dynamic behavior of a single-link rotating flexible arm under large flexural deformation. They used AMM with the first mode approximation to truncated the governing PDE of arm vibrations to a second order nonlinear ODE. We also use AMM method in this paper. According to AMM, the flexible model can be written as

$$v(s,t) = \sum_{n=1}^{N} \phi_n(s) q_n(t), \qquad (10)$$

where $q_n(t)$ is modal coordinates and $\phi_n(s)$ is called shape functions. The eigen functions of linear Euler-Bernoulli beam with clamped-mass boundary condition can be used as the shape functions [11]; these eigen functions are

$$\phi_n = \cos(\beta_n s) - \cosh(\beta_n s) - \frac{\cos(\beta_n l) + \cosh(\beta_n l)}{\sin(\beta_n l) + \sinh(\beta_n l)} (\sin(\beta_n s) - \sinh(\beta_n s)),$$

$$\omega_n = \sqrt{\frac{EI}{\rho}} \beta_n^2, \tag{11}$$

where $\omega_n$ refers to $n$th natural frequency of a linear cantilever beam with a tip mass. Also, $\beta_n$'s can be obtained from the following frequency equation:

$$1 + \cos(\beta_n l)\cosh(\beta_n l) + \alpha \beta_n l [\cos(\beta_n l)\sinh(\beta_n l) - \cosh(\beta_n l)\sin(\beta_n l)] = 0,$$

$$\alpha = \frac{m}{\rho l}. \tag{12}$$

The dynamic behavior of the beam is dominated by the first vibration if the frequency of input forces and input torque are not high. This is the main assumption in this paper. In this case, Eq. (10) can be reduced to $v(s,t) = q_1(t)\phi_1(s)$ or $v(s,t) = q(t)\phi(s)$ for brevity. By substitution this into kinetic and potential energy of the beam, we are able to take the integration in Eqs. (3) and (4) over spatial variable $s$, and thus removing the spatial dependence. The Hamilton principle can now be applied to derive the governing equations of motion. After some calculations, the governing equations of the motion are described by

$$\begin{bmatrix} M_{ff} & \mathbf{M}_{fr} \\ \mathbf{M}_{rf} & \mathbf{M}_{rr} \end{bmatrix} \begin{bmatrix} \ddot{\tilde{q}} \\ \ddot{\mathbf{R}} \end{bmatrix} = \begin{bmatrix} N_f \\ \mathbf{N}_r \end{bmatrix} + \begin{bmatrix} 0 \\ \mathbf{u} \end{bmatrix}, \tag{13}$$

where $M_f$, $N_f$ are scalars, $\mathbf{M}_{rf}$, $\mathbf{M}_{fr}$, $\mathbf{N}_r$ are vectors having three entries, and $\mathbf{M}_{rr}$ is a $3 \times 3$ symmetric matrix. The corresponding expressions for each term are given in Appendix A; also $\tilde{q}$ is the dimensionless elastic mode indicating the deflection at the tip of the beam; furthermore, $\mathbf{R}, \mathbf{u}$ are dimensionless vectors indicating rigid mode and control inputs, respectively. They are defined as follows:

$$\mathbf{R} = \begin{bmatrix} \theta & \tilde{X} & \tilde{Y} \end{bmatrix}^T, \quad \mathbf{u} = \begin{bmatrix} \frac{\tau l}{EI} & \frac{F_x l^2}{EI} & \frac{F_y l^2}{EI} \end{bmatrix}^T. \tag{14}$$

## 3. Trajectory planning

In this section, we introduce trajectory planning techniques with the aim of suppressing the residual vibration of the arm in a rest to rest motion. A trajectory planning technique tries to find an optimal trajectory that creates minimum excitation of flexible mode while the rigid modes move from an initial position to a final position. Search for an optimal trajectory can be carried out for three rigid mode variables simultaneously. Alternatively, we can also search for an optimal trajectory for one of the rigid modes after defining smooth trajectories for the rest. In this way, a cycloid function would be a suitable choice as a smooth trajectory:

$$R_d(t) = (R_f - R_i)\left\{\frac{t}{T_f} - \frac{1}{2\pi}\sin(\frac{2\pi t}{T_f})\right\} + R_i, \qquad (15)$$

where $R_d$ indicate the desired trajectory and should be replaced by $X_d$, $Y_d$, or $\theta_d$. The constants $R_i$ and $R_f$ are desired initial and final positions. The time constant $T_f$ denotes the desired traveling time.

Furthermore, a cost function should be defined whether it is a single-objective optimization or a multi-objective optimization problem. The objective is to minimize the residual vibrations. Thus, the elastic mode equation should be numerically solved for every trajectory. That is,

$$M_{ff}(q, \mathbf{R}_d)\ddot{\tilde{q}} + \mathbf{M}_{fr}(q, \mathbf{R}_d)\ddot{\mathbf{R}}_d = d_f(\tilde{q}, \dot{\tilde{q}}, \mathbf{R}_d, \dot{\mathbf{R}}_d), \qquad (16)$$

where the vector $\mathbf{R}_d$ denotes the desired optimal trajectories. The cost function is defined as

$$\text{cost}(\mathbf{R}_d) = \max_{t > 2T_f}\{\tilde{q}(t) - \bar{\tilde{q}}\}, \qquad (17)$$

where $\bar{\tilde{q}}$ refers to the steady state value of $\tilde{q}$. The cost indicates the amplitude of vibrations in the flexible arm when it reaches a steady state, most often after the time $t = 2T_f$.

The equilibrium equations can be derived by setting all time derivatives to zero. That is,

$$\begin{cases} -\tilde{c}_2\bar{\bar{q}} - 2\tilde{c}_3\bar{\bar{q}}^3 + \lambda_4[2\tilde{c}_4\bar{\bar{q}}sin(\bar{\theta}) - \tilde{c}_5cos(\bar{\theta})] + \lambda_2(2\tilde{c}_1(l)\bar{\bar{q}}sin(\bar{\theta}) - \phi(l)cos(\bar{\theta}))] - \lambda_5^2[2\tilde{c}_{11}\bar{\bar{q}}^3] = 0, \\ \lambda_4[(\tilde{c}_4\bar{\bar{q}}^2 - \frac{1}{2})cos(\bar{\theta}) + \tilde{c}_5\bar{\bar{q}}sin(\bar{\theta}) + \lambda_2(\phi(l)\bar{\bar{q}}sin(\bar{\theta}) + (\tilde{c}_1(l)\bar{\bar{q}}^2 - 1)cos(\bar{\theta}))] + \frac{\bar{\tau}l}{EI} = 0, \\ \bar{F}_x = 0, \\ \bar{F}_y = \lambda_4(1 + \lambda_1 + \lambda_2 + 2\frac{\lambda_3}{\lambda_6^2}), \end{cases} \quad (18)$$

where constants $c_i$ and $\lambda_i$ are in Appendix A. The final value of hub angle $\bar{\theta}$ yields the value of steady state torque $\bar{\tau}$ and subsequently the steady state tip deflection $\bar{\bar{q}}$. It should be noted that the steady state value of the force $\bar{F}_x$ and $\bar{F}_y$ are independent of the hub angle.

In order to generate a trajectory, two types of function are used; that is, spline function or a function generated by an ANN. After that, PSO technique is employed to find the optimal trajectory. These strategies are elaborated in the next subsections.

### 3.1. Spline function

Spline functions have been used for trajectory planning because of its easy use [29]. To construct a spline trajectory, the time interval should be divided into $N+1$ subintervals by $N$ equally distanced midpoints, a discrete function $R_{d,n} = R_d(t_n = T_f/n)$, $n = 1,...,N$ can then be used to generate a spline (see e.g. [46]). The first and second time derivatives of the spline are set to zero (i.e. $\ddot{R}_d = \dot{R}_d = 0$ at $t = 0, T_f$) in order to have a smooth trajectory. Afterwards, Eq. (16) is solved numerically and the cost is calculated. The particle swarm optimization algorithm changes the discrete function $R_{d,n}$ iteratively in such a way that the cost is reduced. This algorithm was introduced by Kennedy and Eberhart [47]. It is a randomized population-based optimization method. A population of candidate solutions, which are also called particles, moves around in a search-space. Each particle is influenced by its own best previous experience and the best experience of all other particles. The swarm eventually moves toward the best solution. The following steps explains this algorithm.

**Step 1:** The position and velocity of the particles are randomly initialized. The position and velocity of *i*th particle are indicated by $\vec{x}_i$ and $\vec{v}_i$, respectively. The position of each particle corresponds to the discrete function $R_{d,n}$ (i.e. $\vec{x}_i = [R_{d,1},...,R_{d,n},...,R_{d,N}]_i$) that is used to generate a spline function.

**Step 2:** Equation (16) is numerically solved and the cost is calculated. This will be used to calculate the best previous position of *i*th particle ($\vec{p}_i$) and the best group position ($\vec{g}$). For the first step, the best previous position is set to initial position.

**Step 3:** The position and velocity of each particle are updated as follows:

$$\vec{v}_i^{k+1} = \chi[\vec{v}_i^k + c_1 r_1^k (\vec{p}_i - \vec{x}_i^k) + c_2 r_2^k (\vec{g} - \vec{x}_i^k)],$$
$$\vec{x}_i^{k+1} = \vec{x}_i^k + \vec{v}_i^{k+1}, \quad (19)$$

where $i = 1, 2, ..., N_{par}$; also, $k$ is the iteration number. The numbers $r_1^k$ and $r_2^k$ are random numbers in the interval [0,1]; they are changed in each iteration. The constants $c_1$ and $c_2$ are two independent constants that influence the rate of convergence of the algorithm. The factor $\chi$ is a constriction factor defined as follows:

$$\chi = \frac{2}{\left|2 - \phi - \sqrt{\phi^2 - 4\phi}\right|}, \quad \phi = c_1 + c_2, \quad \phi > 4. \quad (20)$$

**Step 4:** The cost of new particle positions is calculated. The vectors $\vec{p}_i$ or $\vec{g}$ will be updated if the new positions yield a lower cost.

**Step 5:** Convergence criterion is checked. If the criterion is satisfied, the algorithm will be terminated, and $\vec{g}$ is the optimal solution. Otherwise, the steps 3 to 5 are repeated.

### 3.2. Artificial neural network

In this section, artificial neural networks (ANN) is used to generate an optimal trajectory. The idea was first proposed by Abe [30] in that an ANN is used for trajectory planning of a flexible manipulator with small deformations. The author uses PSO algorithm for training of the ANN. We make use of the same method but for a general planar motion of the large deformable manipulator in previous section. As shown in Figure 2, a three layer ANN is considered to generate a smooth trajectory. It consists of an input layer, a hidden layer with *K* neurons, and an output layer.

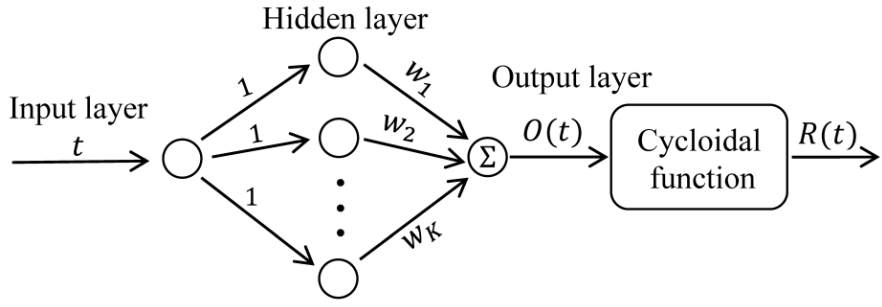

**Fig 2.** Schematic diagram of an ANN.

Sigmoid activation functions are defined for hidden nodes; that is

$$y_k(t) = \frac{1}{1+exp(-a_k(t-b_k))} - \frac{1}{1+exp(a_k b_k)}, \qquad (21)$$

where $a_k$, $k=1,...,K$, are some positive numbers to be optimized by the PSO algorithm. The number $b_k$ is the bias value of $k$th neuron defied as follows

$$b_k = \frac{1-k}{1-K} T_f. \qquad (22)$$

A cycloid activation function is applied to ensure the smoothness of the obtained trajectory; that is,

$$R_d(t) = (R_f - R_i)(O(t) - \frac{1}{2\pi}\sin(2\pi O(t))) + R_i, \qquad (23)$$

where $O(t) = \sum_{k=1}^{K} w_k y_k(t)$. The function $O(t)$ vanishes at the beginning and end of the time interval so that $R_d(0) = R_i$ and $R_d(T_f) = R_f$ are satisfied. The last weight, $w_K$, should therefore be set to

$$w_K = (1 - \sum_{k=1}^{K-1} w_k y(T_f))/y(T_f). \qquad (24)$$

The rest of parameters $a_k$ and $w_k$ except $w_K$ are to be optimized by PSO algorithm. The position of each particle in the algorithm is defined as

$$\vec{x}_i = [a_1, ..., a_K, w_1, ..., w_K]. \qquad (25)$$

## 4. Augmented sliding mode control

In this section, a controller based on sliding mode control law is designed to guide the rigid modes to track the optimal trajectory. The rigid mode dynamic, without uncertainty, can be obtained by Eq. (13) as follows:

$$\ddot{\mathbf{R}} = \mathbf{F}(q, \dot{q}, \dot{\mathbf{R}}) + \mathbf{B}(q, \dot{q}, \dot{\mathbf{R}})u, \qquad (26)$$

where

$$\begin{cases} \mathbf{F} = (\mathbf{M_{rr}} - \dfrac{1}{M_{ff}} \mathbf{M_{rf}} \mathbf{M_{fr}})^{-1} (\mathbf{d_r} - \dfrac{d_f}{M_{ff}} \mathbf{M_{rf}}), \\ \mathbf{B} = (\mathbf{M_{rr}} - \dfrac{1}{M_{ff}} \mathbf{M_{rf}} \mathbf{M_{fr}})^{-1}. \end{cases} \qquad (27)$$

In sliding mode control method, an uncertainty bound in defined. In out example, the uncertainty is as follows:

$$\begin{aligned} \left| F_j - \hat{F}_j \right| &< \eta_j, \quad j = 1, 2, 3, \\ \mathbf{B} &= (\mathbf{I} + \Delta) \hat{\mathbf{B}}, \end{aligned} \qquad (28)$$

where $\hat{\ }$ is used for nominal values. The elements $F_j$, $\eta_j$ are entries of $\mathbf{F}$ and $\hat{\mathbf{F}}$, respectively. The matrix $\mathbf{I}$ is 3×3 identity matrix, and $\Delta$ is 3×3 with entries $\Delta_{ij}$. It is often assumed that $\Delta_{ij}$ satisfies [48]:

$$\Delta_{ij} < D_{ij}, \qquad (29)$$

where $D_{ij} > 0$, and $\|\mathbf{D}\| \leq 1$.

The system described by Eq. (26) is an under-actuated system. For such systems, sliding surfaces cannot solely stabilize the system. The optimal trajectory will therefore help to stabilize the whole system. The following sliding surfaces are suggested

$$S = \dot{E} + KE + \Gamma(\dot{\tilde{q}} + \gamma_4(\tilde{q} - \bar{\tilde{q}})),$$

$$\Gamma = \begin{bmatrix} \gamma_1 \\ \gamma_2 \\ \gamma_3 \end{bmatrix}, \quad E = \begin{bmatrix} e_1 = \theta(t) - \theta_d(t) \\ e_2 = X(t) - X_d(t) \\ e_3 = Y(t) - Y_d(t) \end{bmatrix}, \quad (30)$$

where $\gamma_i > 0$, and $K = diag(k_1, k_2, k_3)$, $k_i > 0$ is a 3×3 diagonal positive-definite matrix. A tangent hyperbolic function is applied to avoid chattering problem [49]:

$$\dot{S} = -A\tanh(S),$$

$$A = diag(A_1, A_2, A_3), \quad \tanh(S) = \begin{bmatrix} \tanh(s_1) \\ \tanh(s_2) \\ \tanh(s_2) \end{bmatrix}, \quad (31)$$

where $A_i > 0$. Combining Eqs. (26), (30) and (31), the control law is as follows:

$$u = \hat{B}^{-1}[\ddot{R}_d - K\dot{E} - \hat{F} - \Gamma(\ddot{\tilde{q}} + \gamma\dot{\tilde{q}}) - A\tanh(S)]. \quad (32)$$

The term $\ddot{\tilde{q}}$ can be dropped and included in the uncertainty [50]. Suitable values for $A_i$ are derived using the following Lyapunov function:

$$V(t) = \frac{1}{2} S^T S. \quad (33)$$

The time derivative of the Lyapunov function should be non-negative so

$$\begin{aligned}\dot{V}(t) &= S^T[\ddot{E} + K\dot{E} + \Gamma(\ddot{\tilde{q}} + \gamma_4\dot{\tilde{q}})] \\ &= S^T[F + Bu - \ddot{R}_d + K\dot{E} + \Gamma(\ddot{\tilde{q}} + \gamma_4\dot{\tilde{q}})] \\ &= S^T[F + B\hat{B}^{-1}[\ddot{R}_d - K\dot{E} - \hat{F} - \Gamma(\ddot{\tilde{q}} + \gamma_4\dot{\tilde{q}}) - A\tanh(S)] - \ddot{R}_d + K\dot{E} + \Gamma(\ddot{\tilde{q}} + \gamma_4\dot{\tilde{q}})] \\ &= S^T[F - \hat{F} - (I + \Delta)A\tanh(S) + \Delta(\ddot{R}_d - K\dot{E} - \hat{F} - \Gamma(\ddot{\tilde{q}} + \gamma_4\dot{\tilde{q}}))] < 0.\end{aligned} \quad (34)$$

An upper-bound for the last term is derived for $s_i > 0$; that is,

$$\eta_i - \frac{\pi}{2}(1-D_{ii})A_i - \frac{\pi}{2}\sum_{j=1}^{3}D_{ij}A_j + \sum_{j=1}^{3}D_{ij}|R_{dj} - k_j\dot{e}_j - \hat{F}_j - \gamma_j(\ddot{\tilde{q}}+\gamma_4\dot{\tilde{q}})|, \tag{35}$$

where $i = 1, 2, 3$. If the above expression is non-negative, the time derivative of Lyapunov function will be negative. This leads to the following condition

$$\frac{\pi}{2}(\delta_{ij}-D_{ii})A_i = \psi_i + \eta_i + \sum_{j=1}^{3}D_{ij}|R_{dj} - k_j\dot{e}_j - \hat{F}_j - \gamma_j(\ddot{\tilde{q}}+\gamma_4\dot{\tilde{q}})|, \tag{36}$$

where $\psi_i > 0$, $i = 1, 2, 3$ and $\delta_{ji}$ is kronecker delta.

Frobenius-Perron theorem is often used to show that Eq. (36) has a unique solution.

**Theorem (Frobenius-Perron)**:

Consider a square matrix A with non-negative elements. The largest real eigenvalue $\lambda_{max}$ of **A** is non-negative. Furthermore, let $(\mathbf{I} - \lambda^{-1}\mathbf{A})y = z$ where all components of the vector $z$ are non-negative. If $\lambda > \lambda_{max}$, then this equation admits a unique solution $y$ whose components $y_i$ are all non-negative.

## 5. Simulations and Conclusions

Simulations presented in this section are carried out in MATLAB/SIMULINK. The following set of parameters are selected

$$\begin{aligned} &l = 2\,m, \quad J = 5\times10^{-3}\,kg.(m/s)^2, \quad EI = 14.58\,N.m, \\ &g = 10\,N/m, \quad m = 0.05\,kg, \quad M = 0.5\,kg, \quad \rho = 0.27\,kg/m, \end{aligned} \tag{37}$$

Initial states are at $\tilde{X}_i = 0$, $\theta_i = -\frac{\pi}{2}$; and the final states are $\tilde{X}_f = 1$, $\theta_f = 0$, $T_f = 2$ *sec*. For these parameters, the first natural frequency of the linear beam and static deflection of the beam at rest are

$$\omega_n = 24.4 \tfrac{rad}{s}, \quad \bar{\bar{q}} = 0.1250,$$
$$\begin{cases} \phi(l) = -1.78, \ c_1(l) = 1.85, \ c_2 = 9.66, \ c_3 = 13.23, \ c_4 = 0.61, \\ c_5 = -0.68, \ c_6 = 0.71, \ c_7 = 0.93, \ c_8 = 0.77, \ c_9 = -1.00, \ c_{10} = -1.48. \end{cases} \quad (38)$$

Furthermore, in the optimization and trajectory planning, the number of particle is in the range 20-30, number of iteration is in the range 70-50, number of point for spline is 5, number of hidden layer neuron is 5. Also, the constants in Eq. (19) are selected in the following ranges

$$\chi \in [0.95 - 0.8], \ \omega \in [0.7 - 0.85], \ c_1 \in [0.7 - 0.9], \ c_2 \in [0.9 - 1.2]. \quad (39)$$

See appendices A.1 and A.2 to know where these numbers appear. To account for uncertainty, we add *5%* uncertainty to each parameter using $\alpha' = \alpha(1 + 0.1 \times sin(20t))$. The numbers $\alpha, \alpha'$ indicate nominal and uncertain parameter, respectively. We should also expect that the performance of the proposed method would improve further if a linear model (i.e. $c_1(s) = c_3 = 0 \rightarrow c_6 = c_7 = c_{10} = 0$) is considered in the simulation. Since natural damping is present in all flexible systems, small residual vibration in the arm would decay eventually. Moreover, the trajectory generated by the ANN yields a better performance compared to the spline trajectories.

Fig 1 illustrates vibration of elastic mode, rotation of arm, slider position and normalized control where cycloid trajectory selected for both rigid modes. The same experiment is done in Fig 2 for spline-based trajectory tracking. Fig 3 shows the results for trajectory generated by the ANN. In Fig 4, 5 and 6, effect of tip mass, gravitational force, and large bending deformation are displayed, respectively.

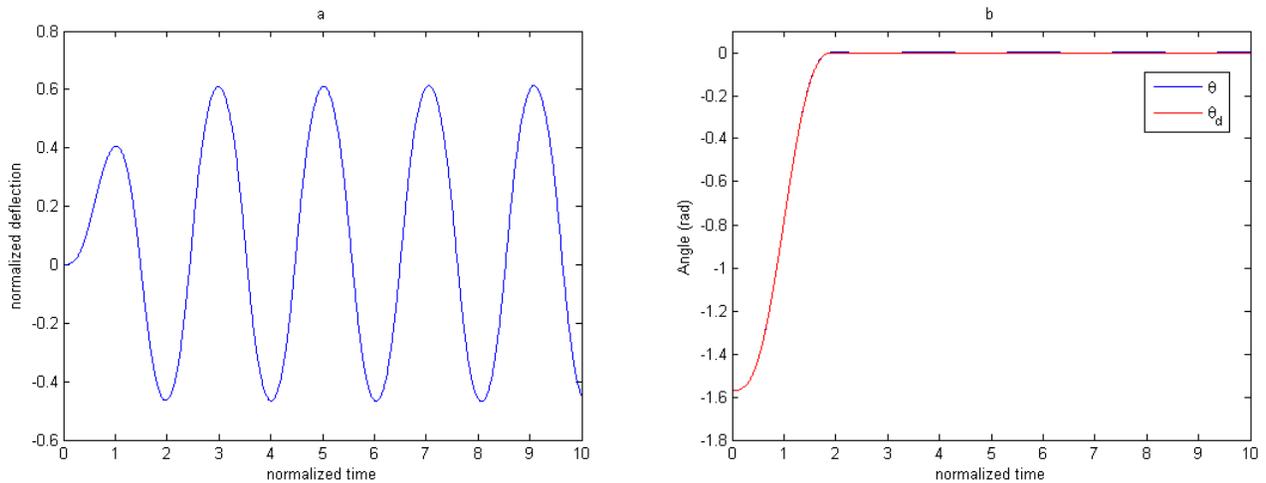

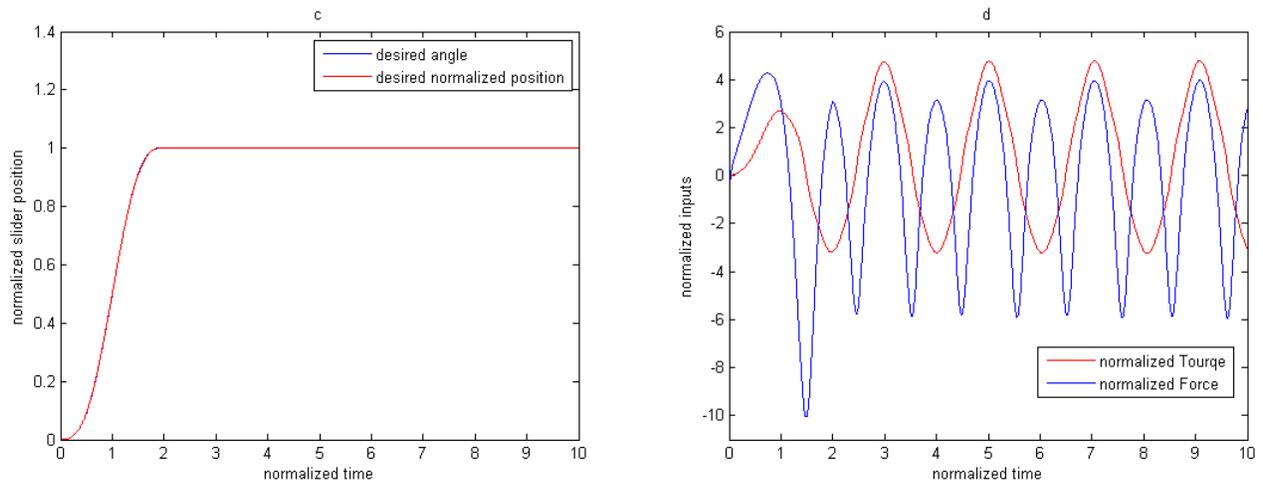

Fig 3. Nonlinear beam model of manipulator with sliding mode control with cycloid trajectory: (a) elastic mode, (b) arm angle, (c) slider position, (d) control input

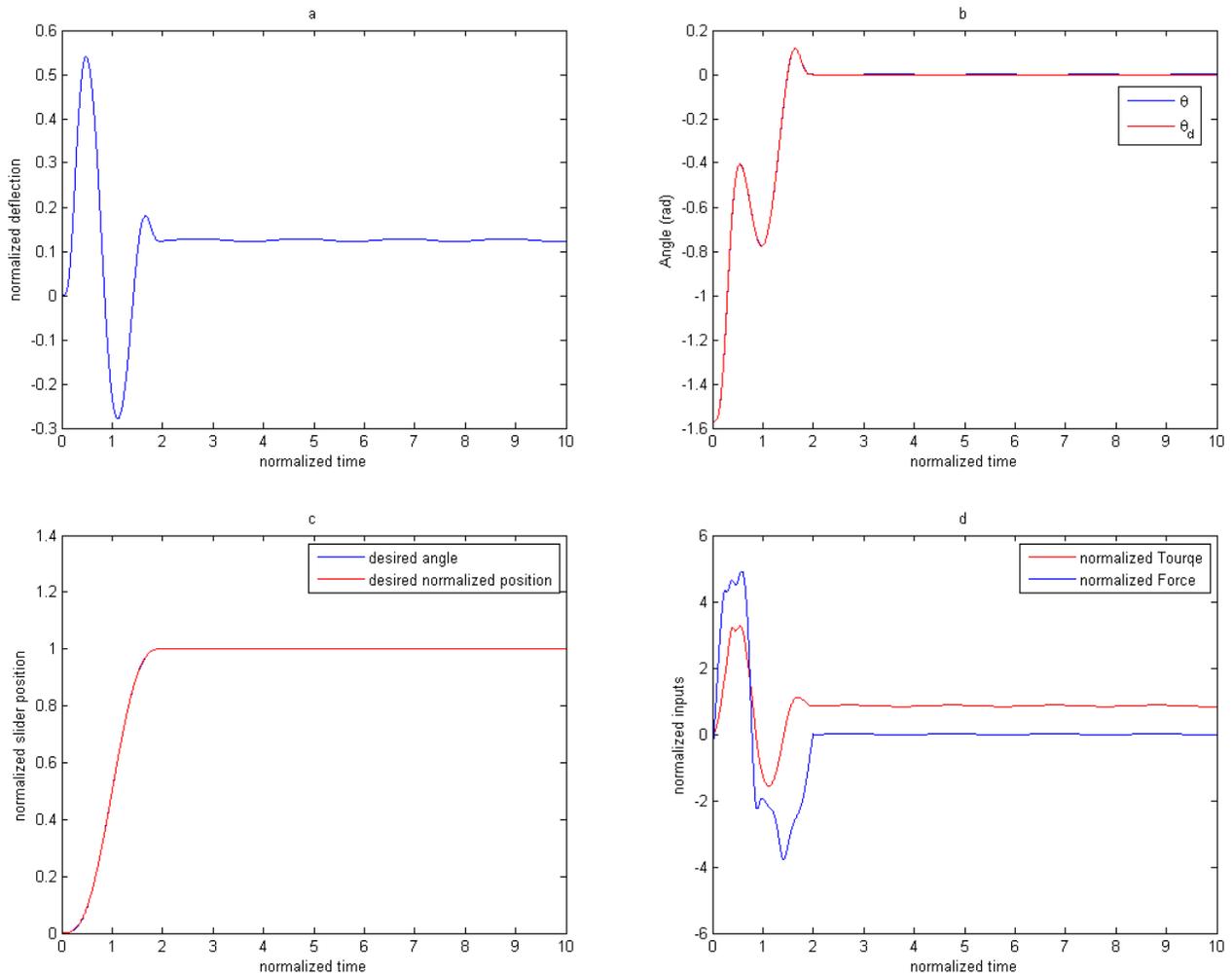

Fig 4. Nonlinear beam model of manipulator with sliding mode control with spline trajectory (ref. A.1.): (a) elastic mode, (b) arm angle, (c) slider position, (d) control input

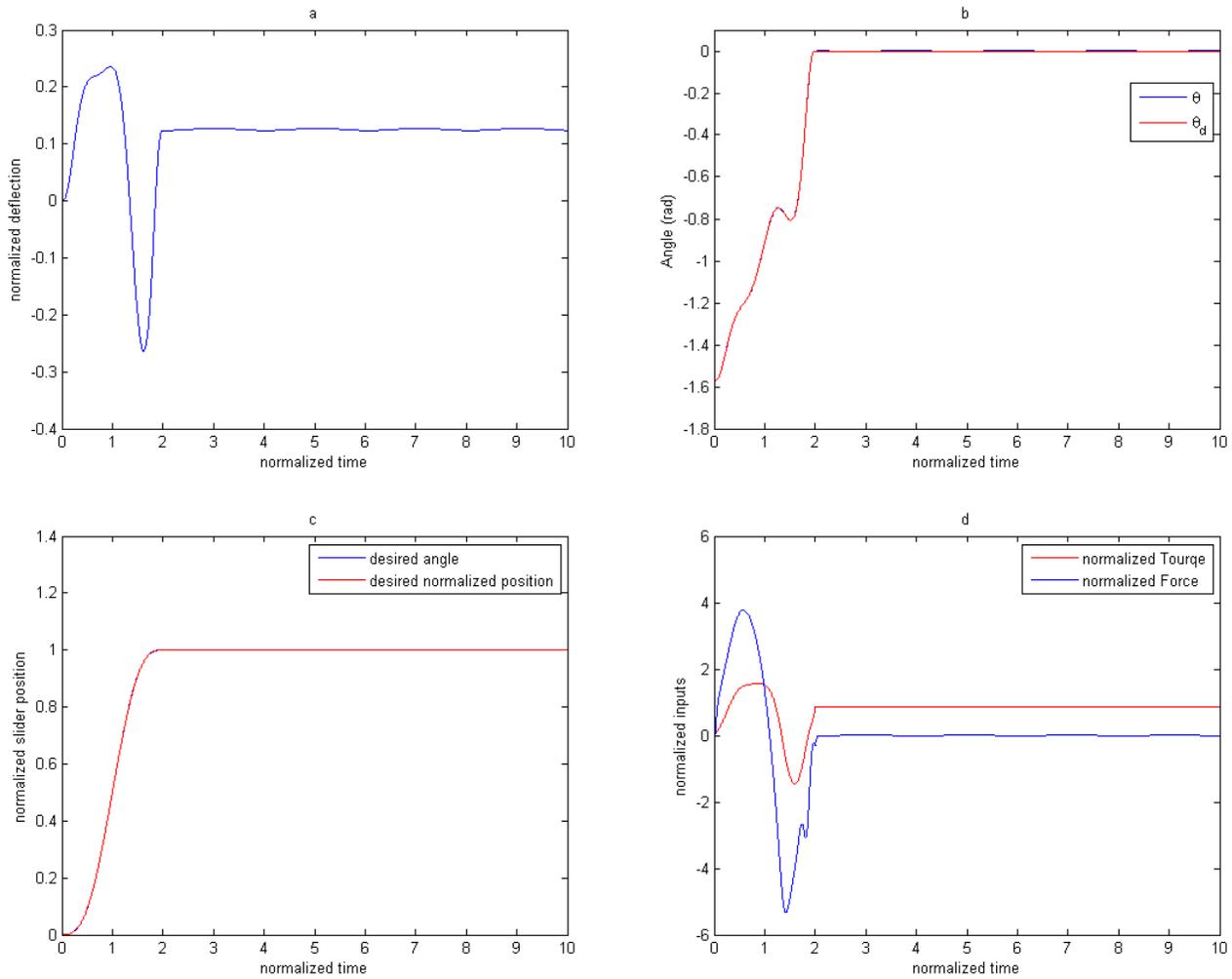

Fig 5. Nonlinear beam model of manipulator with sliding mode control with trajectory generated by the ANN (ref. A.2.): (a) elastic mode, (b) arm angle, (c) slider position, (d) control input

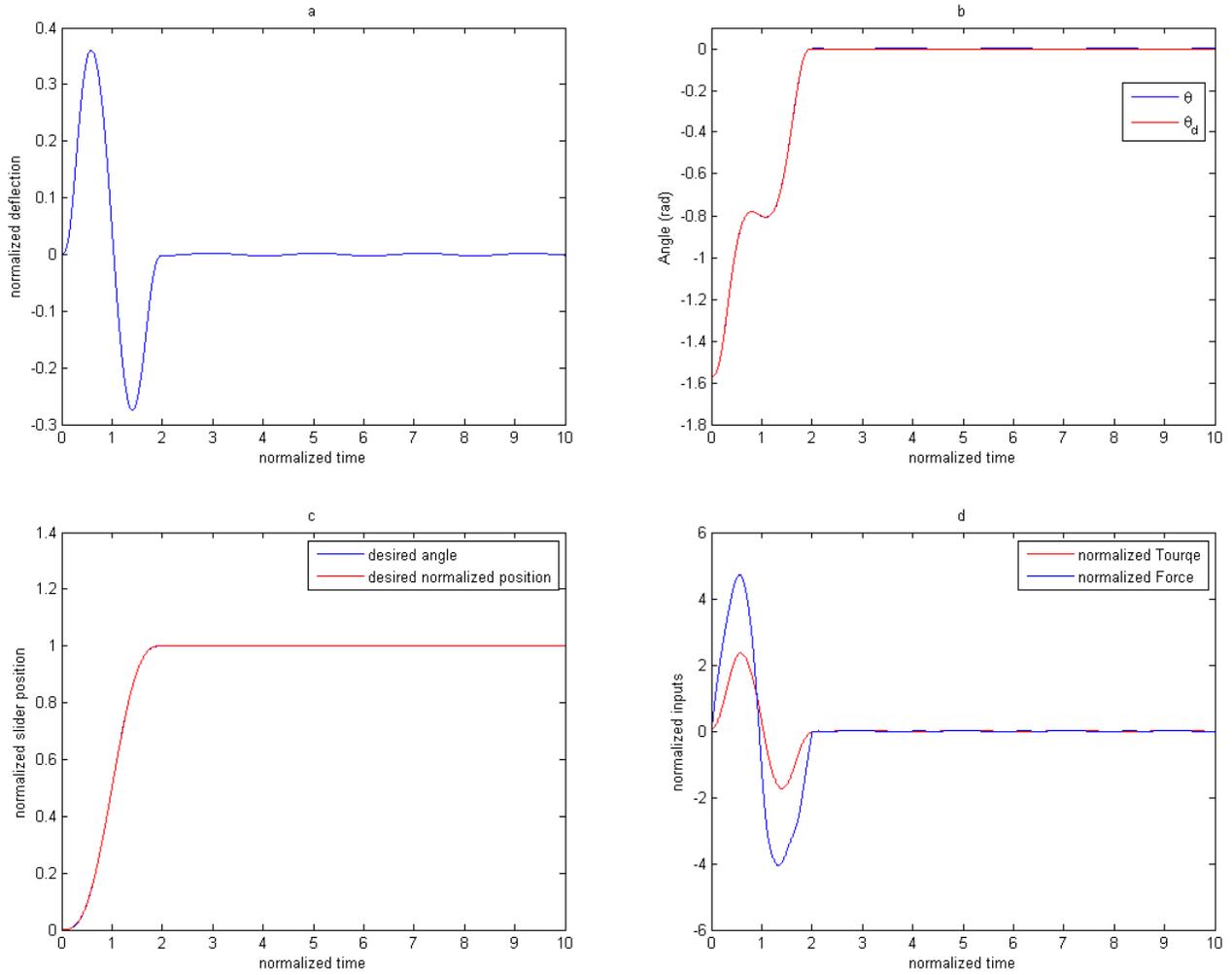

Fig 6. Nonlinear beam model of manipulator with sliding mode for trajectory generated by the ANN in the absence of gravity: (a) elastic mode, (b) arm angle, (c) slider position, (d) control input

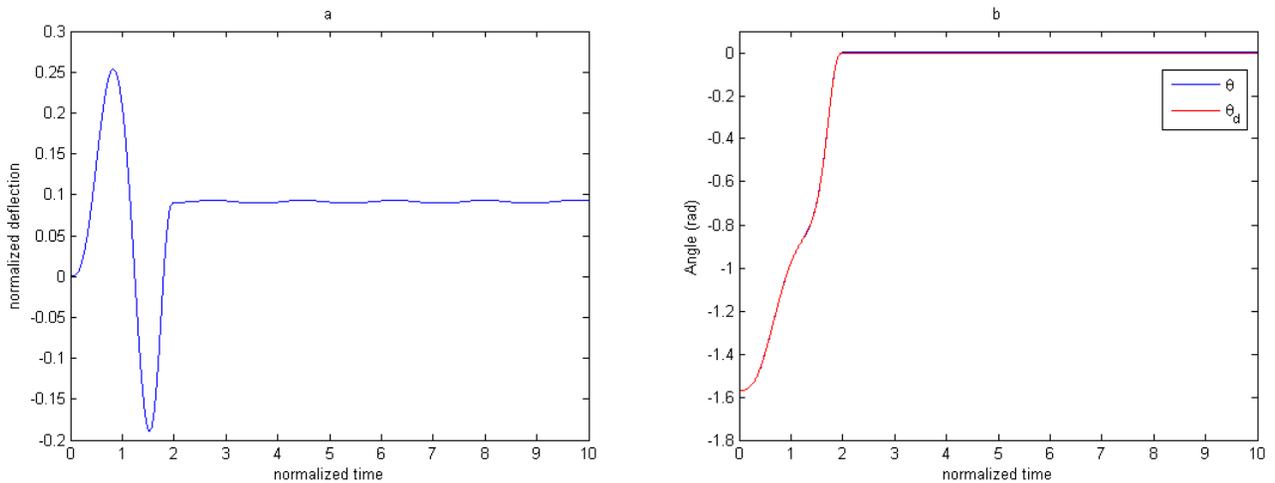

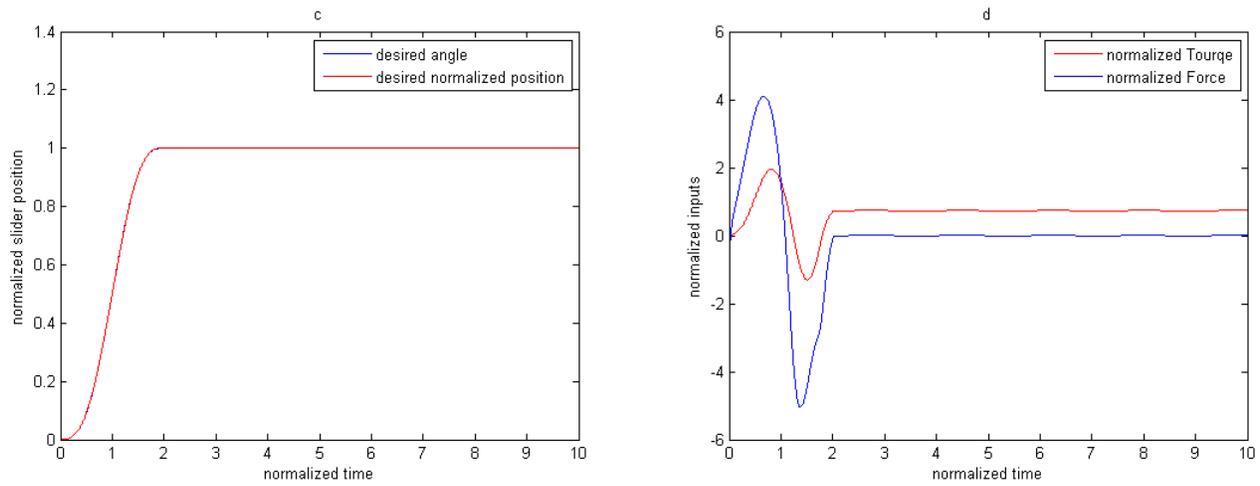

Fig 7. Nonlinear beam model of manipulator with sliding mode control for the trajectory generated by the ANN without a tip mass: (a) elastic mode, (b) arm's angle, (c) slider position, (d) control input

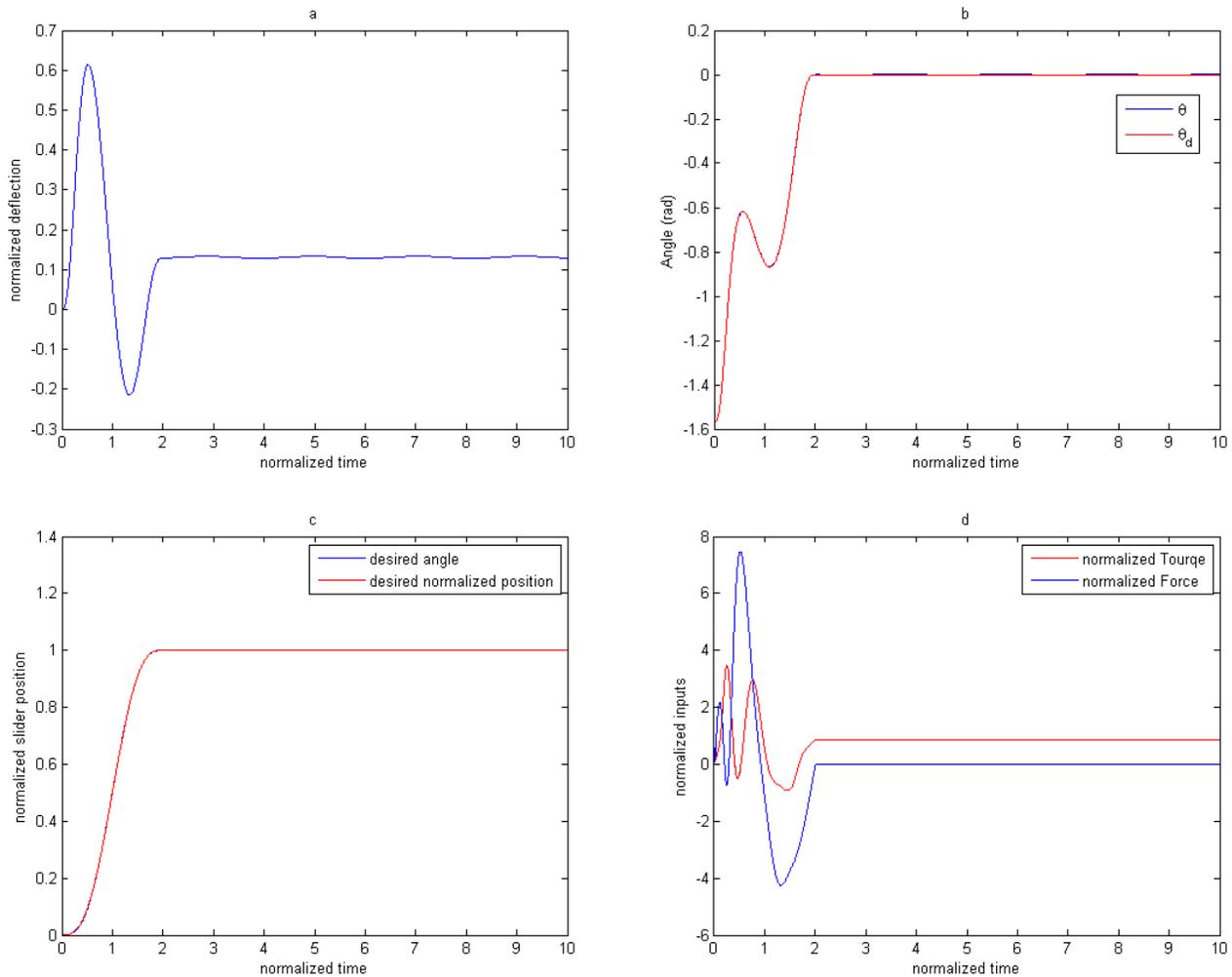

Fig 8. Linear beam model of manipulator with sliding mode control with trajectory generated by the ANN: (a) elastic mode, (b) arm's angle, (c) slider position, (d) control input

## Appendix A:

The nonlinear terms introduced in Eq. (13) are as follows:

$$M_{ff} = \tilde{c}_8 + 4\tilde{c}_6 \tilde{q}^2 + \lambda_2 (\phi(l)^2 + 4\tilde{c}_1(l)^2 \tilde{q}^2), \tag{A.1}$$

$$\mathbf{M}_{rf}(1) = \mathbf{M}_{fr}^T(1) = \frac{1}{2}\tilde{c}_9 + \frac{1}{2}\tilde{c}_{10}\tilde{q}^2 + \lambda_2 \phi(l)(1 + \tilde{c}_1(l)\tilde{q}^2) + \lambda_6(\tilde{c}_5 + \lambda_2), \tag{A.2}$$

$$\mathbf{M}_{rf}(2) = \mathbf{M}_{fr}^T(2) = -(\tilde{c}_5 + \lambda_2 \phi(l))\sin(\theta) - (2\tilde{c}_4 + 2\lambda_2 \tilde{c}_1(l))\tilde{q}\cos(\theta), \tag{A.3}$$

$$\mathbf{M}_{rf}(3) = \mathbf{M}_{fr}^T(3) = (\tilde{c}_5 + \lambda_2 \phi(l))\cos(\theta) - (2\tilde{c}_4 + 2\lambda_2 \tilde{c}_1(l))\tilde{q}\cos(\theta), \tag{A.4}$$

$$\begin{aligned}\mathbf{M}_{rr}(1,1) = &\, c_6 \tilde{q}^4 + (\tilde{c}_8 - \tilde{c}_7)\tilde{q}^2 + \lambda_2 (1 + (\phi(l)^2 - 2\tilde{c}_1(l))\tilde{q}^2 + \tilde{c}_1(l)^2 \tilde{q}^4) \\ &\, + \lambda_3 + \lambda_6 (1 + \lambda_6 + \lambda_2 (\lambda_6 + 2 - 2\tilde{c}_1(l)\tilde{q}^2) - 2\tilde{c}_4 \tilde{q}^2) + \frac{1}{3},\end{aligned} \tag{A.5}$$

$$\mathbf{M}_{rr}(2,2) = \mathbf{M}_{rr}(3,3) = \lambda_1 + \lambda_2 + 1, \tag{A.6}$$

$$\mathbf{M}_{rr}(1,2) = \mathbf{M}_{rr}(2,1) = (-\frac{1}{2} - \lambda_6 (1 + \lambda_2) + \tilde{c}_4 \tilde{q}^2 + \lambda_2 (\tilde{c}_1(l)\tilde{q}^2 - 1))\sin(\theta) - (\tilde{c}_5 + \lambda_2 \phi(l))\tilde{q}\cos(\theta), \tag{A.7}$$

$$\mathbf{M}_{rr}(1,3) = \mathbf{M}_{rr}(3,1) = (\frac{1}{2} + \lambda_6 (1 + \lambda_2) - \tilde{c}_4 \tilde{q}^2 + \lambda_2 (1 - \tilde{c}_1(l)\tilde{q}^2))\cos(\theta) - (\tilde{c}_5 + \lambda_2 \phi(l))\tilde{q}\sin(\theta), \tag{A.8}$$

$$\begin{aligned}N_f = &\, -\tilde{c}_2 \tilde{q} + (2\tilde{c}_6 \dot{\theta}^2 - 2\tilde{c}_3)\tilde{q}^3 - 4c_6 \tilde{q}\dot{\tilde{q}}^2 + (\tilde{c}_8 - \tilde{c}_7)\tilde{q}\dot{\theta}^2 - \tilde{c}_{12}\dot{\theta}^2 \tilde{q} \\ &\, - \lambda_2 [4\tilde{c}_1(l)^2 \tilde{q}\dot{\tilde{q}}^2 - \phi(l)^2 \tilde{q}\dot{\theta}^2 - 2\tilde{c}_1(l)^2 \tilde{q}^3 \dot{\theta}^2 + 2\tilde{c}_1(l)\tilde{q}\dot{\theta}^2] - \lambda_6 [2\dot{\theta}^2 \tilde{q}(\tilde{c}_4 + \lambda_2 \tilde{c}_1(l))] \\ &\, + \lambda_4 [2\tilde{c}_4 \tilde{q}\sin(\theta) - \tilde{c}_5 \cos(\theta) + \lambda_2 (2\tilde{c}_1(l)\tilde{q}\sin(\theta) - \phi(l)\cos(\theta))] - \lambda_5^2 [2\tilde{c}_{11}\tilde{q}^3],\end{aligned} \tag{A.9}$$

$$\begin{aligned}\mathbf{N}_r(1) = &\, -4\tilde{c}_6 \tilde{q}^3 \dot{\tilde{q}}\dot{\theta} + 2(\tilde{c}_7 - \tilde{c}_8)\tilde{q}\dot{\tilde{q}}\dot{\theta} - \tilde{c}_{10}\tilde{q}\dot{\tilde{q}}^2 + 2\tilde{c}_{12}\tilde{q}\dot{\tilde{q}}\dot{\theta} + \lambda_6 [4\tilde{q}\dot{\tilde{q}}\dot{\theta}(\tilde{c}_4 + \lambda_2 \tilde{c}_1(l)] \\ &\, + \lambda_4 [(\tilde{c}_4 \tilde{q}^2 - \frac{1}{2})\cos(\theta) + \tilde{c}_5 \tilde{q}\sin(\theta) + \lambda_2 (\phi(l)\tilde{q}\sin(\theta) + (\tilde{c}_1(l)\tilde{q}^2 - 1)\cos(\theta))] \\ &\, + \lambda_2 [(4\tilde{c}_1(l) - 2\phi(l)^2)\tilde{q}\dot{\tilde{q}}\dot{\theta} - 2\tilde{c}_1(l)\phi(l)\tilde{q}\dot{\tilde{q}}^2 - 4\tilde{c}_1(l)^2 \tilde{q}^3 \dot{\tilde{q}}\dot{\theta}],\end{aligned} \tag{A.10}$$

$$\begin{aligned}\mathbf{N}_r(2) = &\, (\frac{1}{2}\dot{\theta}^2 + 2\tilde{c}_4 \dot{\tilde{q}}^2 - \tilde{c}_4 \tilde{q}^2 \dot{\theta}^2 + 2\tilde{c}_5 \dot{\tilde{q}}\dot{\theta}) \times \cos(\theta) - (\tilde{c}_5 \tilde{q}\dot{\theta}^2 + 4\tilde{c}_4 \tilde{q}\dot{\tilde{q}}\dot{\theta}) \times \sin(\theta) + \lambda_6 [\dot{\theta}^2 \cos(\theta)(1 + \lambda_2)] \\ &\, + \lambda_2 [(\dot{\theta}^2 + 2\tilde{c}_1(l)\dot{\tilde{q}}^2 - \tilde{c}_1(l)\tilde{q}^2 \dot{\theta}^2 + 2\phi(l)\dot{\tilde{q}}\dot{\theta}) \times \cos(\theta) - (\phi(l)\tilde{q}\dot{\theta}^2 + 4\tilde{c}_1(l)\tilde{q}\dot{\tilde{q}}\dot{\theta}) \times \sin(\theta)],\end{aligned} \tag{A.11}$$

$$\mathbf{N_r}(3) = (\frac{1}{2}\dot{\theta}^2 + 2\tilde{c}_4\tilde{q}\dot{\tilde{q}} - \tilde{c}_4\tilde{q}^2\dot{\theta}^2 + 2\tilde{c}_5\dot{\tilde{q}}\dot{\theta}) \times sin(\theta) + (\tilde{c}_5\tilde{q}\dot{\theta}^2 + 4\tilde{c}_4\tilde{q}\dot{\tilde{q}}\dot{\theta}) \times cos(\theta)$$
$$+ \lambda_2[(\dot{\theta}^2 + 2\tilde{c}_1(l)\dot{\tilde{q}}^2 - \tilde{c}_1(l)\tilde{q}^2\dot{\theta}^2 + 2\phi(l)\dot{\tilde{q}}\dot{\theta}) \times sin(\theta) + (\phi(l)\tilde{q}\dot{\theta}^2 + 4\tilde{c}_1(l)\tilde{q}\dot{\tilde{q}}\dot{\theta}) \times cos(\theta)] \quad (A.12)$$
$$- \lambda_4(1 + \lambda_1 + \lambda_2 + \frac{2\lambda_3}{\lambda_6^2}) + \lambda_6[\dot{\theta}^2 sin(\theta)(1 + \lambda_2)].$$

In order to check the validity of calculations, identical expressions were also obtained by MATLAB Symbolic Toolbox [51]. All variables and constants in above expressions are normalized and dimensionless, they are defined as follows:

$$\tilde{t} = t\sqrt{\frac{\rho l^4}{EI}}, \quad \tilde{q} = \frac{q}{l}, \quad \tilde{X} = \frac{X}{l}, \quad (A.13)$$

$$\lambda_1 = \frac{M}{\rho l}, \quad \lambda_2 = \frac{m}{\rho l}, \quad \lambda_3 = \frac{J}{\rho l^3}, \quad \lambda_4 = \frac{\rho g l^3}{EI}, \quad \lambda_5 = \frac{l}{k}, \quad \lambda_6 = \frac{R_H}{l}, \quad (A.14)$$

$$\tilde{c}_1 = lc_1, \quad \tilde{c}_2 = l^3 c_2, \quad \tilde{c}_3 = l^5 c_3, \quad c_4 = \tilde{c}_4, \quad \tilde{c}_5 = \frac{c_5}{l}, \quad \tilde{c}_6 = lc_6,$$
$$\tilde{c}_7 = \frac{c_7}{l}, \quad \tilde{c}_8 = \frac{c_8}{l}, \quad \tilde{c}_9 = \frac{c_9}{l^2}, \quad \tilde{c}_{10} = c_{10}, \quad \tilde{c}_{11} = l^3 c_{11}, \quad \tilde{c}_{12} = \frac{c_{12}}{l}, \quad (A.15)$$

where $\lambda_5$ is also called the slenderness ratio. This is the ratio of the beam length to gyration radius of the cross-sectional area of the beam (i.e., $k = \sqrt{I/A}$). Constants $c_1$ to $c_{10}$ are derived from integration of the spatial functions in the kinetic and potential energies. They are derived as follows

$$c_1(s) = \frac{1}{2}\int_0^s \phi'^2 ds, \quad c_2 = \int_0^l \phi''^2 ds, \quad c_3 = \int_0^l \phi''^2 \phi'^2 ds, \quad c_4 = \int_0^l c_1(s) ds,$$
$$c_5 = \int_0^l \phi ds, \quad c_6 = \int_0^l c_1^2 ds, \quad c_7 = \int_0^l 2sc_1 ds, \quad c_8 = \int_0^l \phi^2 ds, \quad c_9 = \int_0^l 2s\phi ds, \quad (A.15)$$
$$c_{10} = \int_0^l 2c_1\phi ds, \quad c_{11} = \int_0^l [c_1'(s)]^2 ds, \quad c_{12} = \int_0^l [R_H(l-s) + \frac{1}{2}(l^2 - s^2)]\phi'^2 ds.$$

## References

1. Uchiyama, M., et al. *Development of a flexible dual-arm manipulator testbed for space robotics*. in *Intelligent Robots and Systems' 90.'Towards a New Frontier of Applications', Proceedings. IROS'90. IEEE International Workshop on*. 1990. IEEE.

2.	Dwivedy, S.K. and P. Eberhard, *Dynamic analysis of flexible manipulators, a literature review.* Mechanism and Machine Theory, 2006. **41**(7): p. 749-777.
3.	Benosman, M. and G. Le Vey, *Control of flexible manipulators: A survey.* Robotica, 2004. **22**(5): p. 533-545.
4.	Han, S.M., H. Benaroya, and T. Wei, *Dynamics of transversely vibrating beams using four engineering theories.* Journal of Sound and Vibration, 1999. **225**(5): p. 935-988.
5.	Khalil, W. and M. Gautier. *Modeling of mechanical systems with lumped elasticity.* in *Robotics and Automation, 2000. Proceedings. ICRA'00. IEEE International Conference on.* 2000. IEEE.
6.	Book, W.J. *Modeling, design, and control of flexible manipulator arms: A tutorial review.* in *Decision and Control, 1990., Proceedings of the 29th IEEE Conference on.* 1990. IEEE.
7.	Edalatzadeh, M.S. and A. Alasty, *Boundary exponential stabilization of non-classical micro/nano beams subjected to nonlinear distributed forces.* Applied Mathematical Modelling, 2016. **40**(3): p. 2223-2241.
8.	Edalatzadeh, M.S., A. Alasty, and R. Vatankhah, *Admissibility and Exact Observability of Observation Operators for Micro-Beam Model: Time-and Frequency-Domain Approaches.* IEEE Transactions on Automatic Control, 2017. **62**(12): p. 6438-6444.
9.	Edalatzadeh, M.S. and K.A. Morris, *Stability and well-posedness of a nonlinear railway track model.* IEEE control systems letters, 2018. **3**(1): p. 162-167.
10.	Rao, S.S., *Vibration of continuous systems*. 2007: John Wiley & Sons.
11.	Zavodney, L. and A. Nayfeh, *The non-linear response of a slender beam carrying a lumped mass to a principal parametric excitation: theory and experiment.* International Journal of Non-Linear Mechanics, 1989. **24**(2): p. 105-125.
12.	Sathyamoorthy, M., *Nonlinear analysis of beams. Part 1: A Survey of recent advances.* Shock and Vibration Information Center The Shock and Vibration Dig., 1982. **14**(8).
13.	Crespo da Silva, M. and C. Glynn, *Nonlinear flexural-flexural-torsional dynamics of inextensional beams. I. Equations of motion.* Journal of Structural Mechanics, 1978. **6**(4): p. 437-448.
14.	Crespo da Silva, M. and C. Glynn, *Nonlinear flexural-flexural-torsional dynamics of inextensional beams. II. Forced motions.* Journal of Structural Mechanics, 1978. **6**(4): p. 449-461.
15.	Nayfeh, A.H. and P.F. Pai, *Non-linear non-planar parametric responses of an inextensional beam.* International Journal of Non-Linear Mechanics, 1989. **24**(2): p. 139-158.
16.	Nayfeh, A.H. and P.F. Pai, *Linear and nonlinear structural mechanics*. 2008: Wiley. com.
17.	Edalatzadeh, M.S., et al., *Optimal actuator design for vibration control based on LQR performance and shape calculus.* arXiv preprint arXiv:1903.07572, 2019.
18.	Edalatzadeh, M.S. and K.A. Morris, *Optimal actuator design for semilinear systems.* SIAM Journal on Control and Optimization, 2019. **57**(4): p. 2992-3020.
19.	Edalatzadeh, M.S. and K.A. Morris, *Optimal controller and actuator design for nonlinear parabolic systems.* arXiv preprint arXiv:1910.03124, 2019.
20.	Spong, M.W., K. Khorasani, and P.V. Kokotovic, *An integral manifold approach to the feedback control of flexible joint robots.* Robotics and Automation, IEEE Journal of, 1987. **3**(4): p. 291-300.
21.	Siciliano, B. and W.J. Book, *A singular perturbation approach to control of lightweight flexible manipulators.* The International Journal of Robotics Research, 1988. **7**(4): p. 79-90.
22.	Lewis, F.L. and M. Vandegrift. *Flexible robot arm control by a feedback linearization/singular perturbation approach.* in *Robotics and Automation, 1993. Proceedings., 1993 IEEE International Conference on.* 1993. IEEE.
23.	Kokotovic, P., H.K. Khali, and J. O'reilly, *Singular perturbation methods in control: analysis and design*. Vol. 25. 1987: Society for Industrial and Applied Mathematics.
24.	Yin, H., et al., *Theoretical and experimental investigation on decomposed dynamic control for a flexible manipulator based on nonlinearity.* Journal of Vibration and Control, 2013: p. 1077546312474945.
25.	Park, K.-j. and Y.-s. Park, *Fourier-based optimal design of a flexible manipulator path to reduce residual vibration of the endpoint.* Robotica, 1993. **11**(3): p. 263-72.

26. Park, K.-J., *Path design of redundant flexible robot manipulators to reduce residual vibration in the presence of obstacles.* Robotica, 2003. **21**(3): p. 335-340.
27. Park, K.-J., *Flexible robot manipulator path design to reduce the endpoint residual vibration under torque constraints.* Journal of sound and vibration, 2004. **275**(3): p. 1051-1068.
28. Benosman, M., et al., *Rest-to-rest motion for planar multi-link flexible manipulator through backward recursion.* Journal of dynamic systems, measurement, and control, 2004. **126**(1): p. 115-123.
29. Abe, A., *Trajectory planning for residual vibration suppression of a two-link rigid-flexible manipulator considering large deformation.* Mechanism and Machine Theory, 2009. **44**(9): p. 1627-1639.
30. Abe, A., *Trajectory planning for flexible Cartesian robot manipulator by using artificial neural network: numerical simulation and experimental verification.* Robotica, 2011. **29**(5): p. 797-804.
31. Damaren, C. and I. Sharf, *Simulation of flexible-link manipulators with inertial and geometric nonlinearities.* Journal of dynamic systems, measurement, and control, 1995. **117**(1): p. 74-87.
32. Boyer, F. and W. Khalil, *Kinematic model of a multi-beam structure undergoing large elastic displacements and rotations.: Part one: model of an isolated beam.* Mechanism and machine theory, 1999. **34**(2): p. 205-222.
33. Boyer, F. and W. Khalil, *Kinematic model of a multi-beam structure undergoing large elastic displacements and rotations. Part two: kinematic model of an open chain.* Mechanism and machine theory, 1999. **34**(2): p. 223-242.
34. Boyer, F., N. Glandais, and W. Khalil, *Flexible multibody dynamics based on a non-linear Euler–Bernoulli kinematics.* International journal for numerical methods in engineering, 2002. **54**(1): p. 27-59.
35. Martins, J., M.A. Botto, and J.S. da Costa, *Modeling of flexible beams for robotic manipulators.* Multibody System Dynamics, 2002. **7**(1): p. 79-100.
36. Yang, J., L. Jiang, and D. Chen, *Dynamic modelling and control of a rotating Euler–Bernoulli beam.* Journal of sound and vibration, 2004. **274**(3): p. 863-875.
37. Xue, X. and J. Tang, *Vibration control of nonlinear rotating beam using piezoelectric actuator and sliding mode approach.* Journal of Vibration and Control, 2008. **14**(6): p. 885-908.
38. Al-Bedoor, B. and M. Hamdan, *Geometrically non-linear dynamic model of a rotating flexible arm.* Journal of sound and vibration, 2001. **240**(1): p. 59-72.
39. Fazel, M.R., M.M. Moghaddam, and J. Poshtan, *Application of GDQ method in nonlinear analysis of a flexible manipulator undergoing large deformation.* Proceedings of the Institution of Mechanical Engineers, Part C: Journal of Mechanical Engineering Science, 2013. **227**(12): p. 2671-2685.
40. Lagnese, J. and G. Leugering, *Uniform stabilization of a nonlinear beam by nonlinear boundary feedback.* Journal of Differential Equations, 1991. **91**(2): p. 355-388.
41. Bayo, E., *A finite-element approach to control the end-point motion of a single-link flexible robot.* Journal of Robotic Systems, 1987. **4**(1): p. 63-75.
42. Bayo, E. *Computed torque for the position control of open-chain flexible robots.* in *Robotics and Automation, 1988. Proceedings., 1988 IEEE International Conference on*. 1988. IEEE.
43. Vatankhah, R., et al., *Boundary Stabilization of Non-Classical Micro-Scale Beams.* Applied Mathematical Modelling, 2013.
44. Theodore, R.J. and A. Ghosal, *Comparison of the assumed modes and finite element models for flexible multilink manipulators.* The International journal of robotics research, 1995. **14**(2): p. 91-111.
45. Al-Bedoor, B., A. El-Sinawi, and M. Hamdan, *Non-linear dynamic model of an inextensible rotating flexible arm supported on a flexible base.* Journal of sound and vibration, 2002. **251**(5): p. 767-781.
46. Lin, C., P. Chang, and J. Luh, *Formulation and optimization of cubic polynomial joint trajectories for industrial robots.* Automatic Control, IEEE Transactions on, 1983. **28**(12): p. 1066-1074.
47. Kennedy, J. and R. Eberhart. *Particle swarm optimization.* in *Proceedings of IEEE international conference on neural networks.* 1995. Perth, Australia.
48. Park, K.B. and T. Tsuji, *Terminal sliding mode control of second-order nonlinear uncertain systems.* International Journal of Robust and Nonlinear Control, 1999. **9**(11): p. 769-780.

49. Song, G. and H. Gu, *Active vibration suppression of a smart flexible beam using a sliding mode based controller.* Journal of Vibration and Control, 2007. **13**(8): p. 1095-1107.
50. Wilson, D.G., et al., *Augmented sliding mode control for flexible link manipulators.* Journal of Intelligent and Robotic Systems, 2002. **34**(4): p. 415-430.
51. Toolbox, S.M., *Matlab.* Mathworks Inc, 1993.